\pdfoutput=1  

\documentclass{isprs} 
\usepackage{setspace}
\usepackage{url} 
\usepackage{geometry} 
\usepackage[labelsep=period]{caption}  
\usepackage[british]{babel} 
\usepackage[hang]{footmisc}
\usepackage[dvipsnames]{xcolor}
\usepackage{colortbl}
\usepackage{subcaption}
\usepackage{float}
\usepackage{geometry} 
\usepackage[hidelinks]{hyperref}
\usepackage{todonotes}
\usepackage[labelsep=period]{caption}  

\usepackage{natbib}
\usepackage{multicol}
\usepackage{graphicx}
\usepackage{todonotes}
\usepackage{amsfonts}
\usepackage{listings}
\definecolor{dkgreen}{rgb}{0,0.6,0}
\definecolor{gray}{rgb}{0.5,0.5,0.5}
\definecolor{mauve}{rgb}{0.58,0,0.82}
\definecolor{lavender}{rgb}{0.9, 0.9, 0.98}
\usepackage{multirow}

\definecolor{lightgray}{gray}{0.92}
\definecolor{headergray}{gray}{0.85}
\usepackage{booktabs}

\usepackage{pgfplots}
\pgfplotsset{compat=1.18}
\usepackage{pgf-pie}

\usepackage{sansmath}

\usepackage{amsmath}

\usetikzlibrary{positioning,
  arrows.meta,         
  matrix,              
  ext.node-families,   
  ext.positioning-plus,
  ext.paths.ortho      
}
\tikzset{
  rows/.style 2 args={
    /utils/temp/.style={row ##1/.append style={nodes={#2}}},
    /utils/temp/.list={#1}},
  columns/.style 2 args={
    /utils/temp/.style={column ##1/.append style={nodes={#2}}},
    /utils/temp/.list={#1}}}
\makeatletter
\tikzset{
  left delimiter/.style 2 args={append after command={
    \tikz@delimiter{south east}{south west}
    {every delimiter,every left delimiter,#2}{south}{north}{#1}{.}{\pgf@y}}}}
\makeatother

\usepackage[]{todonotes}

\usepackage{booktabs}
\usepackage[normalem]{ulem}
\usepackage[locale=UK]{siunitx}
\usepackage[ruled,vlined,linesnumbered]{algorithm2e}

\begin{document}

\title{SVI2LoD3: Agent-Driven Reconstruction of LoD3 Façade Openings in Semantic 3D City Models from Volunteered Street View Imagery using Large Language and Visual Models}
\date{}



\author{
 Elmehdi Kanna\textsuperscript{*}, Lukas Arzoumanidis\textsuperscript{*}, Huynh Duc An Son Nguyen, Youness Dehbi}

\address{Computational Methods Lab, HafenCity University, Hamburg, Germany - \\ \{elmehdi.kanna, lukas.arzoumanidis, son.nguyen, youness.dehbi\}@hcu-hamburg.de}

\abstract{
This paper presents an end-to-end, agent-driven pipeline for the LoD3 reconstruction of façade openings in 3D city models, producing directly usable CityGML-conform outputs. In contrast to existing approaches that rely on supervised semantic segmentation and therefore require large amounts of manually annotated training data, the proposed method employs a zero-shot segmentation strategy. This substantially reduces the annotation effort while still achieving strong performance in our benchmark on the eTRIMS dataset. A further key contribution is the enforcement of correct partonomic hierarchies, thereby producing CityGML-conform LoD3 building models. Beyond the reconstruction pipeline itself, this work also introduces a novel evaluation metric for façade reconstruction, termed \textit{Facade Feature Distance} (FFD). Unlike conventional metrics such as mIoU or FRDS, which assess similarity primarily through pixel-wise overlap, FFD measures distance in a high-level feature space derived from a vision transformer. In doing so, it captures both semantic correctness and architectural layout, providing a more suitable assessment of façade reconstruction quality. The proposed pipeline and evaluation strategy together offer a practical and scalable contribution toward the automated generation and analysis of semantically enriched 3D city models. The developed code is published at: \url{https://github.com/hcu-cml/citydb-SVI2LoD3-ai}.

}


\maketitle

{\renewcommand\thefootnote{\fnsymbol{footnote}}
\footnotetext[1]{These authors contributed equally to this work.}
}

\sloppy

\section{Introduction}\label{MANUSCRIPT}

Advances in urban analysis and simulation tasks, such as modeling of urban temperature and wind flow, building energy demand estimation, and simulations of interactions between autonomous vehicles and Internet of Things (IoT) infrastructures, are increasingly dependent on highly detailed and semantically rich 3D city models \citep{Kolbe2021}. These applications require not only accurate geometric representations of the urban environment but also rich semantic information.

Currently, the majority of publicly available semantic 3D city models remain at LoD2 \citep{isprs-archives-XLVIII-4-2024-493-2024, isprs-annals-IV-4-W1-51-2016}. While LoD2 models typically include differentiated roof structures and generalized building geometries, they remain limited in terms of detailed representations for façades \citep{GROGER201212}. Consequently, fine-grained façade characteristics, such as material properties, façade composition, and the precise geometry and placement of openings, are missing. This limitation is primarily driven by the complexity of acquiring and reconstructing such detailed information at scale. Addressing this gap by enabling city-wide LoD3 reconstruction therefore requires advanced methods capable of capturing the heterogeneity of architectural styles and structural typologies \citep{PANG2022102859}.

Moreover, reconstruction processes typically rely on observations that are inherently noisy and incomplete, such as airborne or terrestrial LiDAR point clouds and street-view imagery. These observations are affected by measurement errors, occlusions, varying illumination conditions, and perspective distortions \citep{Kolbe2021}. 
Since airborne and terrestrial LiDAR point cloud coverage is unavailable for most urban regions, volunteered street-view imagery (SVI) can serve as an alternative data source due to its broader availability. This wider coverage is largely attributable to the considerably less demanding acquisition and handling of camera images for contributors, compared to large point clouds from different vendors, which typically require additional processing.

\begin{figure*}[ht!]
    \centering
    \includegraphics[width=\textwidth]{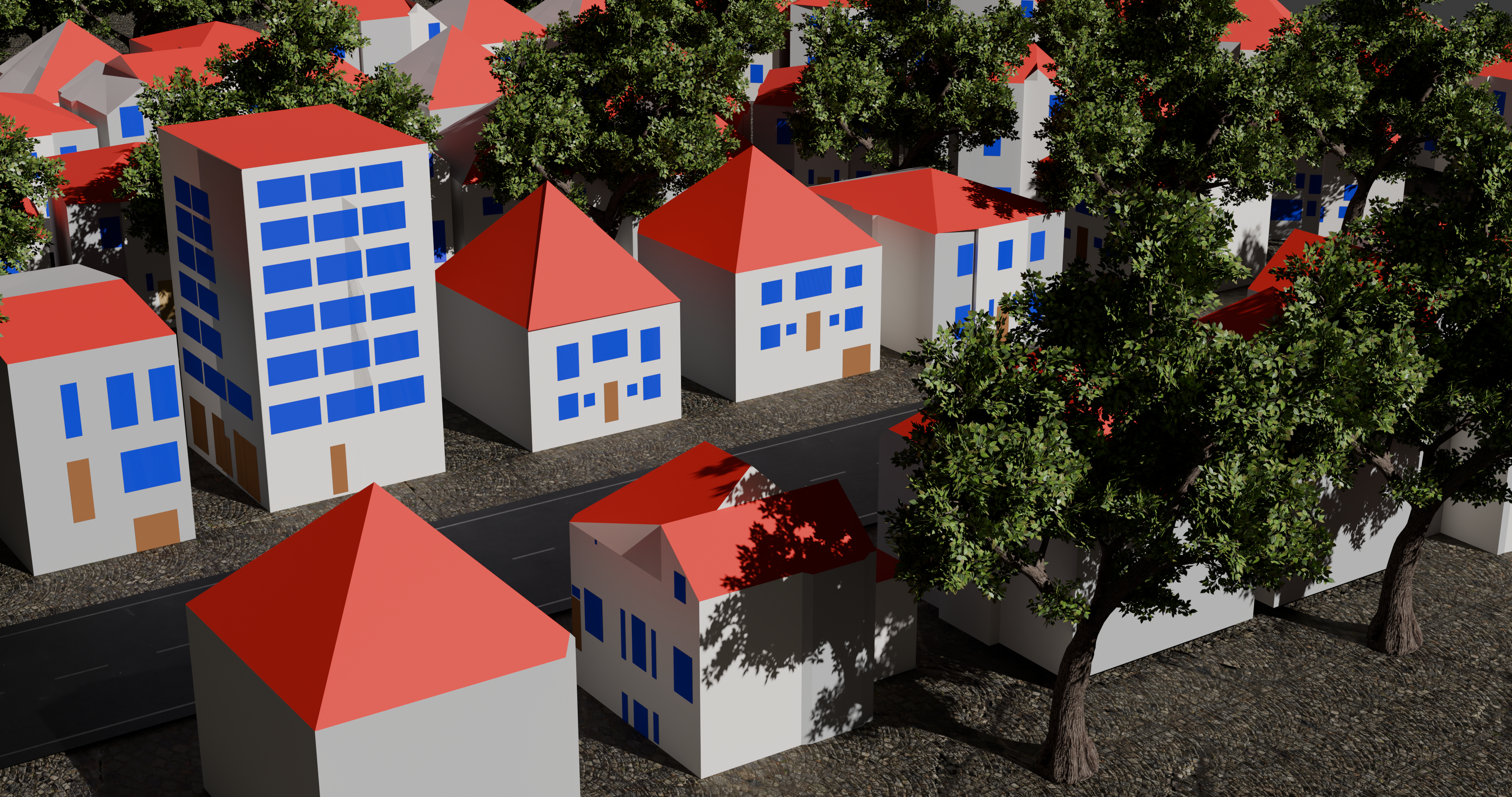}
    \caption{Reconstructed façade openings for LoD3 building models in a selected exemplary neighborhood in Hamburg, Germany, visualized in Blender\textsuperscript{\textcopyright}. Blue rectangles indicate reconstructed windows, while brown rectangles indicate reconstructed doors.}
    \label{fig:3d_model}
\end{figure*}

As a result, this work proposes a novel agent-driven approach for the reconstruction of façade openings from SVI in semantic 3D city models using large-scale foundation models, as illustrated in Figure~\ref{fig:3d_model}. Specifically, our approach leverages the high-level reasoning capabilities of Large Language Models (LLMs) and zero-shot image segmentation capabilities of Large Vision Models (LVMs) within an agent-driven framework to guide and coordinate the reconstruction process. To ensure syntactic as well as semantic, geometric, and topological correctness of the reconstructed 3D city models in accordance with the CityGML 2.0 LoD3 schema, we developed algorithms that support the agent in two distinct stages of the reconstruction process, which we refer to as \textit{expert-designed programs}. 


To the best of our knowledge, this approach is among the first to combine expert-designed programs into an agent-driven reconstruction approach. By coupling these complementary capabilities, the proposed approach is able to address the substantial architectural variability of building façades, including diverse arrangements and geometries of façade openings which has historically posed a significant challenge for automated reconstruction approaches and has limited the generalizability of many existing methods, as will be pointed out in Section \ref{redlated}. 
In addition, this work introduces a novel evaluation metric for façade openings in reconstructed LoD3 models that relies on a high-level feature distance instead of pixel-wise comparison strategies commonly adopted in previous approaches.

Our main contributions include:
\begin{itemize}
    \item an end-to-end approach for reconstructing CityGML-conformant LoD3 building models from volunteered SVI,
    \item the use of LVM for zero-shot segmentation of façade openings, eliminating the need for task-specific training data and enabling direct application to different cities and architecturally heterogeneous façades,
    \item an agent-driven framework that leverages the high-level reasoning capabilities of large language models for 2D-to-3D projection in combination with expert-designed algorithms, and
    \item a novel evaluation metric that focuses on semantic and topological accuracy, in contrast to state-of-the-art metrics that assess only semantic accuracy.
\end{itemize}




\section{Related Work}\label{redlated}

In recent years, the reconstruction of façade openings has attracted increasing attention, with numerous deep learning-based approaches proposed that utilize observational data such as airborne or terrestrial LiDAR-derived point clouds, street-view imagery (SVI), or airborne RGB imagery. The following subsections review several of the most recent and influential approaches in this domain and highlight the key distinctions between these methods and the approach proposed in this work.

\subsection{LiDAR-Based Façade Reconstruction}

Over the past years, numerous machine learning-based approaches have been developed for the 3D reconstruction of buildings using terrestrial and airborne LiDAR point clouds. These methods typically exploit the geometric richness of point cloud data to infer building structures and façade elements.

Using terrestrial LiDAR point clouds, \citet{article_Dehbi} employed Support Vector Machines (SVMs) in combination with Statistical Relational Learning (SRL) through Markov Logic Networks (MLNs) to learn grammar-based rules for the reconstruction of 3D building structures. This approach integrates probabilistic reasoning with structural constraints to infer building components from point cloud observations. Similarly, a method proposed by \citet{rs8090737} combines multiple geometric algorithms within a rule-based framework to reconstruct building façades from terrestrial LiDAR point cloud data. In this framework, geometric feature extraction and predefined architectural rules are jointly utilized to derive façade structures and opening configurations. More recently, \citet{10208594} introduced an approach aimed at improving the accuracy of semantic LoD3 building reconstruction. Their method enhances façade-level semantic 3D segmentation by incorporating knowledge of laser scanning physics together with prior information derived from existing 3D building models to probabilistically identify model conflicts. 

\subsection{Image-Based Façade Reconstruction}

Using Structure-from-Motion (SfM) and deep-learning-based segmentation techniques, \citet{PANTOJAROSERO2022104430} developed a pipeline for the automatic reconstruction of LoD3 building models of free-standing structures, with a particular focus on detecting and reconstructing façade openings. \citet{PANG2022102859} proposed a method for reconstructing 3D building models from a single street-view image using image-to-mesh reconstruction techniques. Their approach leverages street-view imagery to enhance the level of detail of commonly available block models (LoD1). In the same context, \citep{WANG202490} propose an automated framework for semantic building-model reconstruction at an urban scale. Specifically, they introduce a façade layout graph model to represent the geometric and topological relationships of semantic entities on building façades, thereby enabling the inference of structural completeness and the reconstruction of semantic façade models. Rather than using street-view imagery, their approach exploits mesh textures that are often available for LoD2 models and uses deep learning-based computer vision methods to extract semantic information from these textures in order to enrich the existing LoD2 building models. \citet{11147463} present a method for reconstructing LoD3 building models by combining low-detail 3D building models with panoramic street-level imagery. Their results show that low-detail building models can provide suitable planar reference surfaces for the orthorectification of panoramic images. In addition, they demonstrate that applying semantic segmentation to accurately textured low-detail façade surfaces preserves key properties required for LoD3 reconstruction, including georeferencing consistency, watertight geometry, and a compact low-polygon representation. Recently, \citet{MA2026106842} presented a method for reconstructing façade openings in 3D building models by integrating street-view imagery. The approach introduces a mathematically derived method for estimating unknown intrinsic camera parameters, enabling metric 2D-to-3D projection without relying on multi-view imagery or pre-existing depth information. Furthermore, a supervised semantic segmentation model is employed to detect façades in an example study area in Amsterdam, allowing detailed façade openings to be measured and pixel coordinates to be transformed into spatial coordinates.

\section{Methodology}\label{meth}

To enable the reconstruction of façade openings from street-view imagery (SVI) and their integration into existing  LoD2 building models for the generation of LoD3 models, we employ an agent-driven framework that combines the high-level reasoning capabilities of a Large Language Model (LLM) with the zero-shot image segmentation capabilities of a Large Vision Model (LVM). Within this framework, the agent coordinates the reconstruction workflow by invoking a set of expert-designed programs designed for geometric processing and model integration.

\begin{figure*}[ht!]
    \centering
    \includegraphics[width=0.945\linewidth]{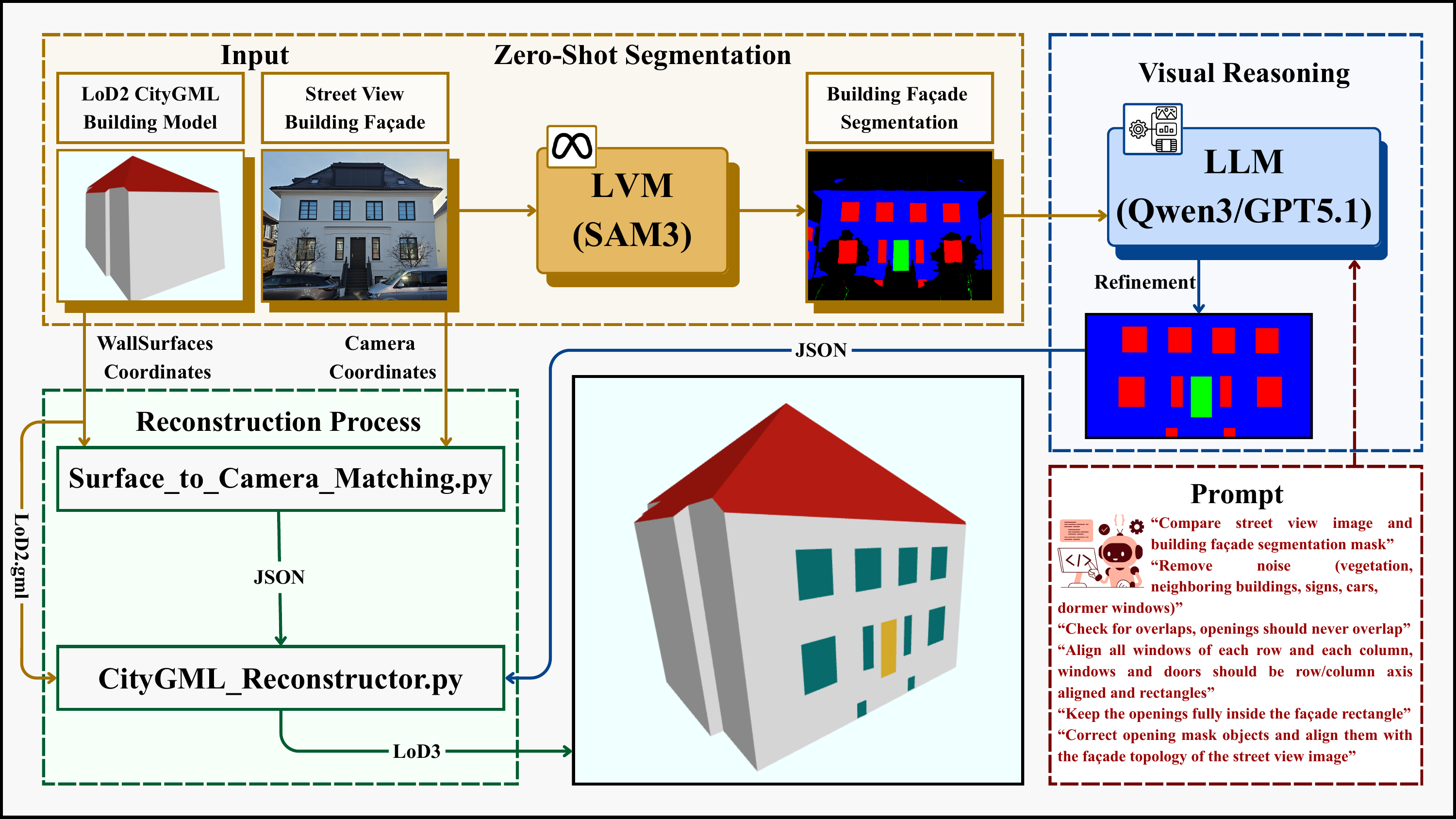}
    \caption{Overview of our agent-driven approach for reconstructing façade openings in valid LoD3 CityGML-conform models.}
    \label{fig:methodology}
\end{figure*}

For façade segmentation from SVI, our approach utilizes an LVM because of the substantial variability in façade design, including differences in architectural style, materiality, and ornamentation. 
Within the agent framework, we evaluate and compare the performance of GPT‑5.1\footnote{\url{https://openai.com/index/gpt-5-1/}} and Qwen3‑VL 30B Instruct\footnote{\url{https://qwen.ai/blog?id=99f0335c4ad9ff6153e517418d48535ab6d8afef&from=research.latest-advancements-list}}
for guiding the LoD3 reconstruction process leveraging their high-level reasoning capabilities~\citep{XU2025101370}. Specifically, these models orchestrate the execution of expert-designed programs that extract façade openings from the segmentation masks obtained from SAM~3 and integrate them into existing LoD2 building models, as illustrated in Figure~\ref{fig:methodology}. 

The comparison between different LLMs as backbone for our agent framework is relevant from a practical deployment perspective. While GPT-5.1 is a commercial model associated with usage costs, Qwen3-VL is available as an open-source alternative. In our case, access to the latter is facilitated through the ChatAI system\footnote{\url{https://docs.hpc.gwdg.de/services/saia/index.html}} \citep{doosthosseini2024chataiseamlessslurmnative}, which provides access to large-scale models hosted on a high-performance computing infrastructure, allowing experimentation with open models in a controlled environment while avoiding additional usage costs.

To improve the LLM's reliability when performing geometrical and mathematical operations during the reconstruction process, our agent is designed to invoke expert-designed programs tailored to specific reconstruction tasks, as highlighted in Figure~\ref{fig:methodology}. These include, for example, façade surface-to-camera image matching based on camera location, as well as the extraction of façade openings from segmented masks and their integration into an LoD2 building model. 

To evaluate our approach, we consider three of the four principal error dimensions. According to the OGC standard definition\footnote{\url{https://www.ogc.org/standards/citygml/}} \citep{Biljecki2016Errors, GROGER201212}, these dimensions are geometry, semantics, topology, and completeness. In this work, we only focus on the semantic, topological, and completeness dimensions, which we assess using a set of state-of-the-art evaluation metrics and a newly defined metric that jointly captures semantic and topological quality within a single score.


\subsection{Zero-Shot Façade Composition Segmentation}

To address the challenge of identifying façade openings in SVI across diverse urban environments characterized by substantial variation in façade design, material composition, and architectural ornamentation, we employ SAM~3. SAM~3 is a LVM that achieves strong predictive performance in object detection and segmentation tasks, including on previously unseen object categories, without requiring task-specific fine-tuning or labeled training data~\citep{carion2025sam3segmentconcepts}.
This zero-shot capability removes the dependence on annotated datasets, whose production is both resource-intensive and time-consuming, and facilitates straightforward transfer of our approach to other urban contexts where Mapillary\footnote{\url{https://www.mapillary.com/}} imagery is available. 

SAM~3 supports text-conditioned segmentation, whereby natural language prompts describing target objects are provided to guide mask generation. Through prompting, the model was instructed to segment windows, doors and façade surfaces within each image, as illustrated in Figure~\ref{fig:methodology}. For every input image, SAM~3 generated multi-class segmentation masks corresponding to regions consistent with the provided textual prompts.

\subsection{LLM-Driven Reasoning for Façade Shape Completion}

After semantic segmentation of the façade image using SAM~3, the approximate geometry of the façade surface is derived from the class-specific color encoding of the predicted segmentation mask. More specifically, façade elements are extracted based on their RGB values, with façades being represented in blue, windows in red, and doors in green.

A refinement step is needed because the segmentation mask obtained from the SVI contains the façade openings required for LoD3 reconstruction, yet it is often degraded by noise, occlusions, and small spurious artifacts due to the inherently imperfect nature of SVI data \citep{HOU2022103094, BILJECKI2021104217}, as illustrated in Figure~\ref{fig:methodology}. To improve the quality and topology of the extracted façade-opening geometries, we prompt the agent to leverage the high-level reasoning capabilities of the underlying LLM for shape completion and reprojection on the semantically segmented mask predicted by SAM~3. Furthermore, in order to preserve genuine openings that are only partially visible because of occlusions caused by vegetation or parked vehicles, the agent retains candidate openings that either have their centroid within the façade geometry or touch the façade boundary.

Moreover, the agent is instructed to remove sources of noise, including vegetation, neighboring buildings, traffic signs, vehicles and dormer windows, as shown in Figure~\ref{fig:methodology}. An additional prompt enforces a consistency check for overlapping façade openings, as such overlaps are not geometrically plausible in the reconstructed model. Finally, to regularize and complete the geometry of façade openings, the agent is prompted to impose structural constraints such that windows and doors are represented as axis-aligned rectangles, while openings within the same horizontal and vertical arrangements are aligned along common rows and columns. These assumptions reflect the geometric regularity typically exhibited by real-world building façades tested in this study.

\subsection{Façade Surface-To-Camera Image Matching}\label{facade_matching}

In order to associate the façade visible in the SVI with the corresponding façade of the LoD2 model to be reconstructed, we developed an algorithm that enables the agent to identify the correct LoD2 façade surface from the camera perspective. For the sake of clarity and reproducibility, Algorithm~\ref{alg:facade_matching_staged} is provided as a reference to guide the reader through the methodological steps described in the following paragraphs.

First, the original camera parameters and positions of the SVI are transformed into the same Coordinate Reference System (CRS) as the LoD2 CityGML model (cf. Algo.~\ref{alg:facade_matching_staged} line 2). To determine which building façade faces the camera, we adopt the assumption that the nearest observable façade is also the façade facing the camera. In CityGML, a façade may consist of multiple \texttt{WallSurface} elements, which can be represented as a set $W$ of \texttt{WallSurfaces} $ws_{1}, \dots, ws_{n}$. Additionally, we define a set $C$ of \texttt{WallSurface} centroids $c_{1}, \dots, c_{n}$. If a \texttt{WallSurface} has an area of $4\,\mathrm{m}^2$ or less, we treat it as a distinct building component, such as a garage, and disregard it as a matchable façade.

\begin{figure}[ht!]
    \centering
    \includegraphics[width=\linewidth]{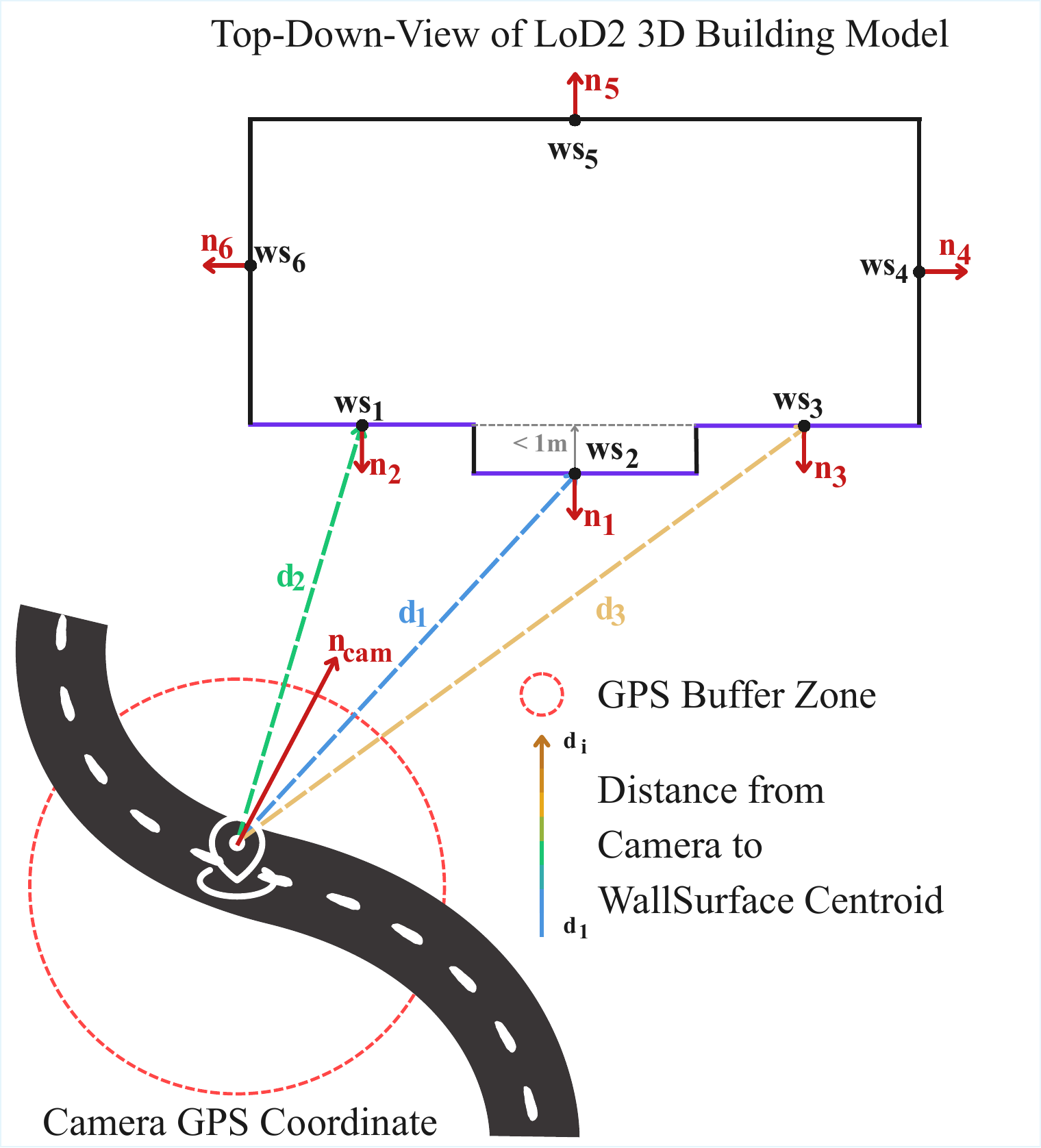}
    \caption{Illustration of our approach for façade surface-to-camera image matching, as described in Algorithm~\ref{alg:facade_matching_staged}. }
    \label{fig:placeholder}
\end{figure}

To identify all façades observable in the camera image, we compute the camera viewing normal vector $n_{\mathrm{cam}}$ corresponding to the image vanishing direction (cf. Algo.~\ref{alg:facade_matching_staged} line 3). For each \texttt{WallSurface} $ws_{i}$, we compute its outward normal vector $n_{i}$ in order to exclude \texttt{WallSurfaces} that are not visible in the camera image. This filtering is performed by comparing each $n_{i}$ with the previously computed camera normal vector $n_{\mathrm{cam}}$ using a photogrammetric visibility criterion based on plane-ray intersection and normal alignment, as illustrated in Figure~\ref{fig:placeholder} (cf. Algo.~\ref{alg:facade_matching_staged} line 5-10). If two or more \texttt{WallSurface} normals yield the same matching quality with respect to $n_{\mathrm{cam}}$, we identify the \texttt{WallSurfaces} that jointly constitute the full camera-facing façade by additionally evaluating their Euclidean distance to the camera position and ranking them accordingly (cf. Algo.~\ref{alg:facade_matching_staged} line 12-20).

The Euclidean distance $d_{i}$ between the camera position, represented by the coordinates $(x_{cam}, y_{cam})$, and the centroid $c_{i}$, represented by the coordinates $(x_{i}, y_{i})$, is computed as follows:
\begin{equation}
\forall c_i \in C:\quad d_i = \sqrt{(x_{i} - x_{cam})^2 + (y_{i} - y_{cam})^2}.
\end{equation}

Because the LoD2 models in our dataset frequently contain small geometric extrusions, e.g., bay windows, we additionally verify whether the closest matching \texttt{WallSurface} and the remaining matching \texttt{WallSurfaces} exhibit a perpendicular separation of at most $1\,\mathrm{m}$. This threshold was chosen, as most observed extrusions fall within this range \citep{handle:20.500.11811/5097}. \texttt{WallSurfaces} whose perpendicular distance is less than $1\,\mathrm{m}$ are retained. The remaining \texttt{WallSurfaces} are then considered to jointly form the true camera-facing façade of the LoD2 model (cf. Algo.~\ref{alg:facade_matching_staged} line 20-26).

Figure~\ref{fig:placeholder} illustrates an example in which the \texttt{WallSurfaces} walls $ws_{1}$, $ws_{2}$, and $ws_{3}$ are forming the true camera-facing façade of the LoD2 model while \texttt{WallSurfaces} $ws_{6}$, $ws_{5}$, and $ws_{4}$ are excluded from the image-to-façade matching process because their normal vectors $n_{6}$, $n_{5}$, $n_{4}$ are not aligned with $n_{\mathrm{cam}}$.

As illustrated in Figure~\ref{fig:methodology}, the matched \texttt{WallSurfaces} are exported as JSON elements containing their CityGML \texttt{wall\_id} and geometric information describing the physical dimensions of each segment (cf. Algo.~\ref{alg:facade_matching_staged} line 28). This is needed to distribute the façade openings derived from the shape-completed segmentation mask in a geometrically and topologically consistent manner when a façade consists of multiple \texttt{WallSurfaces}.


\begin{algorithm}[t]
\caption{Selection of the camera-facing façade from an LoD2 building model}
\label{alg:facade_matching_staged}
\DontPrintSemicolon
\SetKwInOut{Input}{Input}
\SetKwInOut{Output}{Output}

\Input{Camera parameters $\mathcal{C}$, camera position $p_{\mathrm{cam}}$, \texttt{WallSurfaces} set $W = \{ws_1,\dots,ws_n\}$}
\Output{Matched façade $F_{\mathrm{match}}$}

\BlankLine
\textbf{Coordinate transformation and view direction estimation}\;
Transform $\mathcal{C}$ and $p_{\mathrm{cam}}$ to the LoD2 coordinate reference system\;
Compute camera normal vector $n_{\mathrm{cam}}$\;

\BlankLine
\textbf{Visibility and geometric filtering}\;
$W_{\mathrm{obs}} \gets \emptyset$\;
\ForEach{$ws_i \in W$}{
    Compute \texttt{WallSurface} normal $n_i$\;
    Compute \texttt{WallSurface} area $ws_i^{area}$\;
    \If{$ws_i$ satisfies the visibility criterion with respect to $n_{\mathrm{cam}}$ \textbf{and} $ws_i^{area} > 4\,\mathrm{m}^2$} {
        $W_{\mathrm{obs}} \gets W_{\mathrm{obs}} \cup \{ws_i\}$\;
    }
}

\BlankLine
\textbf{Candidate ranking}\;
$W_{\mathrm{cand}} \gets \emptyset$\;
Determine the highest normal-consistency score in $W_{\mathrm{obs}}$\;
\ForEach{$ws_i \in W_{\mathrm{obs}}$}{
    \If{$ws_i$ attains the highest normal-consistency score}{
        Compute centroid $c_i$\;
        Compute $d_i \gets \|p_{\mathrm{cam}} - c_i\|$\;
        $W_{\mathrm{cand}} \gets W_{\mathrm{cand}} \cup \{(ws_i, d_i)\}$\;
    }
}
Sort $W_{\mathrm{cand}}$ by ascending distance\;
Select the nearest candidate $ws_{\mathrm{min}}$\;

\BlankLine
\textbf{Façade aggregation}\;
$F_{\mathrm{match}} \gets \{ws_{\mathrm{min}}\}$\;
\ForEach{$ws_j \in W_{\mathrm{cand}} \setminus \{ws_{\mathrm{min}}\}$}{
    Compute perpendicular distance $\delta_{\perp}(ws_{\mathrm{min}}, ws_j)$\;
    \If{$\delta_{\perp}(ws_{\mathrm{min}}, ws_j) < 1\,\mathrm{m}$}{
        $F_{\mathrm{match}} \gets F_{\mathrm{match}} \cup \{ws_j\}$\;
    }
}

\BlankLine
\textbf{Export}\;
Export $F_{\mathrm{match}}$ with CityGML \texttt{wall\_id} and semantic dimensions\;
\Return{$F_{\mathrm{match}}$}\;
\end{algorithm}

\subsection{LoD3 CityGML Reconstruction and Validation}\label{file_recon}

As an outcome of the reasoning process, the agent produces a JSON file containing a list of façade-opening geometries, specifically windows and doors, and the façade surface geometry together with their corresponding coordinates extracted from the predicted façade segmentation mask after shape completion. We match the façade surface geometry with the previously matched wall surface geomery, ensuring correct partonomy and scaling of openings.

In cases where a façade is represented by multiple \texttt{WallSurface} elements in the LoD2 model, the segmentation mask is subdivided proportionally to the widths of the respective \texttt{WallSurface} elements in the LoD2 geometry, thereby preserving the geometric arrangement and topological consistency of the detected openings.

Subsequently, the LoD3 model is reconstructed by incorporating the detected and matched façade openings into the building model, as highlighted in Figure~\ref{fig:methodology}. The identified windows and doors are encoded as explicit CityGML elements and inserted directly into the XML structure of the CityGML document. The resulting XML representation is then validated using the CityGML schema validation framework\footnote{\url{https://github.com/tudelft3d/CityGML-schema-validation}}. Finally, the validated output is written as a compliant LoD3 CityGML 2.0 document.

\subsection{Benchmarking Dataset \& Evaluation Metrics}

To benchmark our approach, we use the eTRIMS dataset introduced by \citet{korc-forstner-tr09-etrims}. The dataset contains street-view images together with corresponding semantic annotations covering eight semantic classes, namely: sky, façade, window, door, vegetation, car, road and pavement. This dataset was selected because it captures a wide range of façade images representing different architectural styles and cities across European urban environments, where we conducted our test.

\begin{figure}[b!]
  \centering
    \begin{subfigure}{0.32\columnwidth}
  \includegraphics[width=\textwidth]{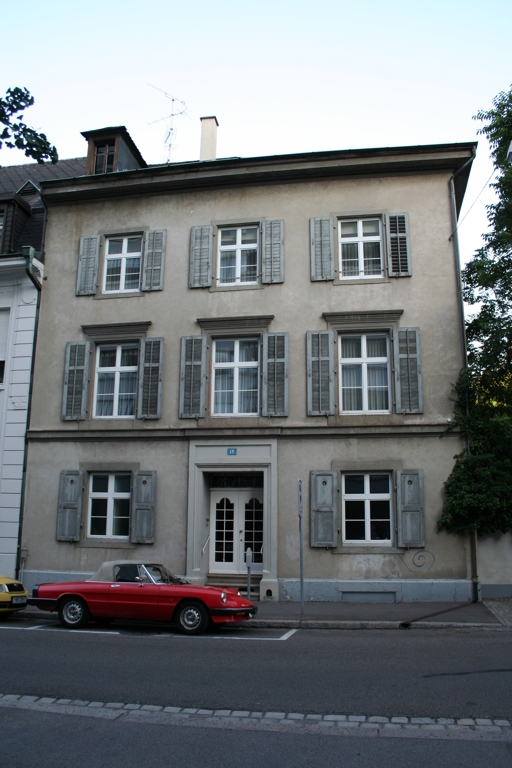}
  \caption{street-view building image}
  \end{subfigure}
  \begin{subfigure}{0.32\columnwidth}
  \includegraphics[width=\textwidth]{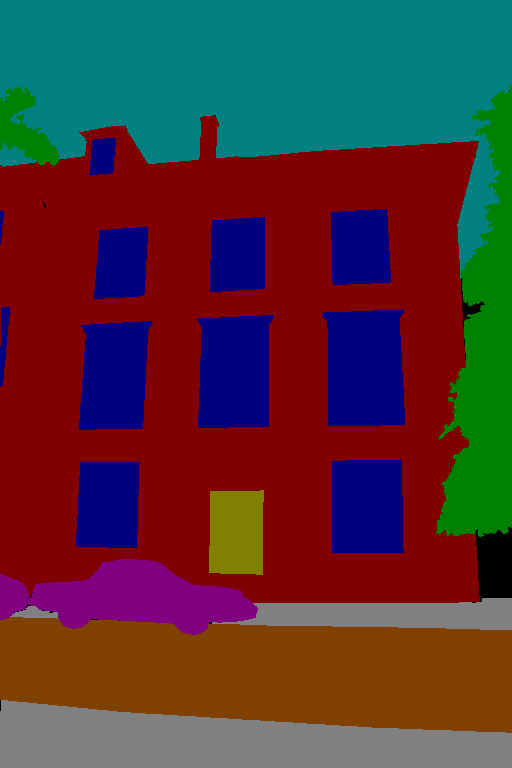}
  \caption{eTRIMS ground truth}
  \end{subfigure}
  \begin{subfigure}{0.32\columnwidth}
  \includegraphics[width=\textwidth]{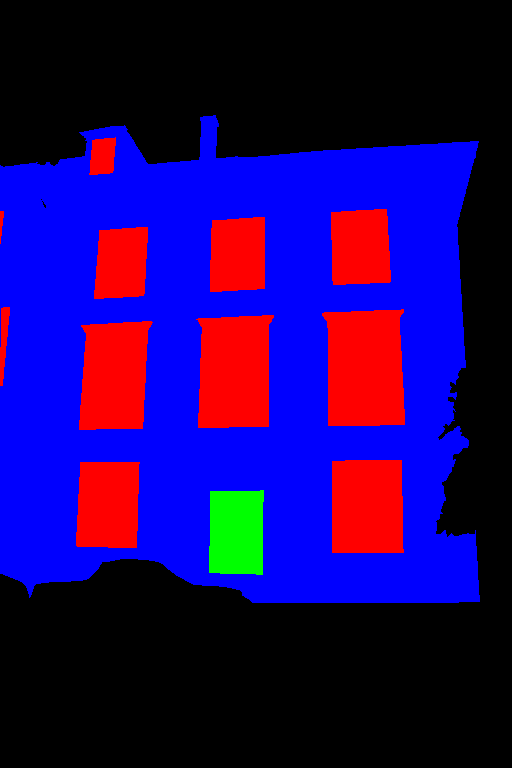}
  \caption{adapted ground truth}
  \end{subfigure} 
\begin{subfigure}{0.32\columnwidth} 
  \includegraphics[width=\textwidth]{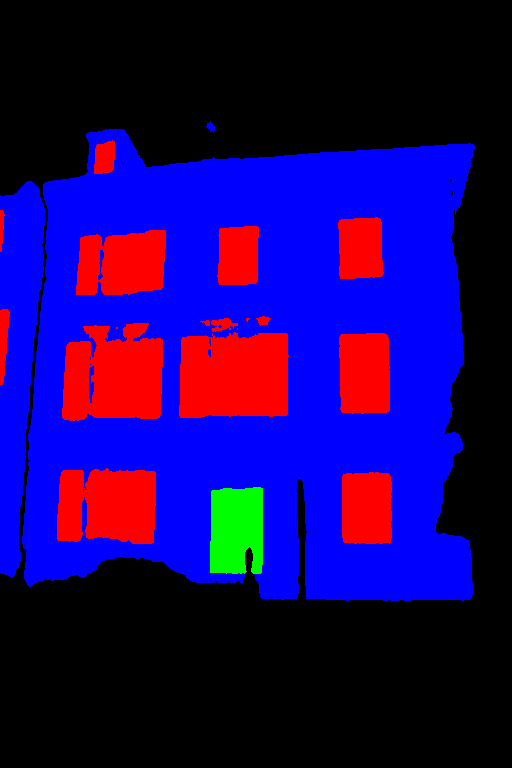} 
  \caption{segmentation mask}
  \end{subfigure}  
  \begin{subfigure}{0.32\columnwidth} 
  \includegraphics[width=\textwidth]{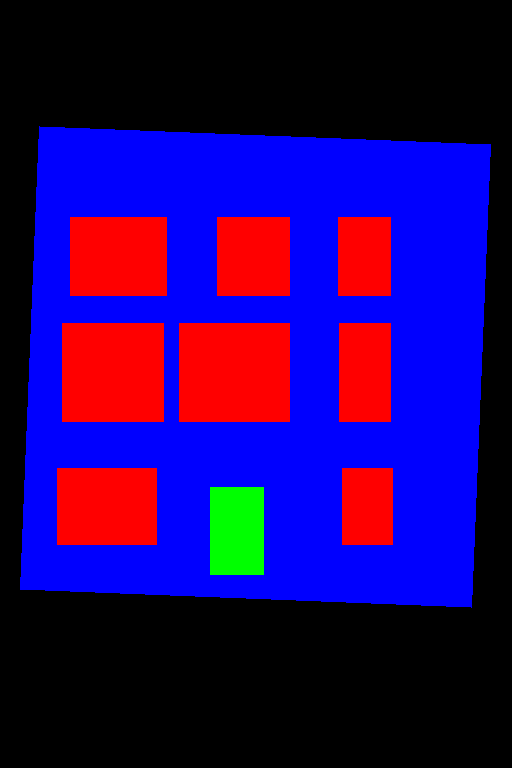} 
  \caption{shape completion}
  \end{subfigure}
  \caption{Original ground truth mask from eTRIMS dataset (b), our ground truth adaptation (c), semantic segmentation using SAM~3 (d) and LLM-driven refinement results (e). Façades represented in blue, windows in red, and doors in green.}
  \label{fig:gt-normalisation}
\end{figure}

Prior to evaluation, classes not relevant to the façade reconstruction task, such as trees, cars, and road and pavement surfaces, were removed from the annotations, as illustrated in Figure~\ref{fig:gt-normalisation}. In addition, the color encoding of the remaining classes, specifically windows, doors, and building façades, was adapted to match the class color scheme produced by SAM~3 in order to enable consistent evaluation. To evaluate our reconstructed LoD3 models, we selected two state-of-the-art metrics for the evaluation of reconstructed façade openings in 3D building models: mean Intersection over Union (mIoU) \citep{WANG202490} and the Facade Reprojection Dice Score (FRDS) \citep{MA2026106842, PANTOJAROSERO2022104430}. Both metrics evaluate the reconstructed LoD3 models purely based on 2D planar geometry extracted from images. However, they neglect the semantic and topological relationships that are critical for successful reconstruction and the generation of valid, CityGML-conform models.

For each class $i$, IoU measures the overlap between the predicted segmentation and the ground-truth annotation. It is defined as the ratio between correctly predicted pixels of a class (true positives) and the union of predicted and ground-truth pixels for that class, including false positives and false negatives. The mIoU is then computed by averaging the IoU values across all $K$ classes:
\begin{equation}
\mathrm{IoU_i} = \frac{\mathrm{TP_i}}{\mathrm{TP_i + FP_i + FN_i}}, \quad \mathrm{mIoU} = \frac{\mathrm{1}}{\mathrm{K}}\sum_{\mathrm{i=1}}^{\mathrm{K}} \mathrm{IoU_i}.
\end{equation}
In the context of façade image segmentation, mIoU quantifies how accurately the model segments façade elements such as windows, doors, and façade surfaces. A higher mIoU indicates better agreement between predicted and annotated regions, reflecting more precise extraction of façade openings.

Building upon the Dice Score, the FRDS compares ground truth masks with the re-projection of modeled façades derived from terrestrial LiDAR point clouds \citep{PANTOJAROSERO2022104430}. This metric assesses both the precision of the surface reconstruction and the layout of the façade openings. Similar to the mIoU, a perfect reconstruction where the re-projected façade overlaps exactly with the ground truth yields an FRDS of 1.0. Following \citet{PANTOJAROSERO2022104430}, the FRDS is defined as:
\begin{equation}
\mathrm{FRDS} =
\frac{2\,\mathrm{TP}}
{2\,\mathrm{TP} + \mathrm{FP} + \mathrm{FN}}.
\end{equation}
In our approach, the LLM backbone of our agent refines and re-projects the segmentation masks derived from SAM~3, followed by a shape completion step. We consider this process equivalent to the re-projection of modeled façades derived from LiDAR reconstructions. This equivalency holds because our agent extracts the façade openings using expert-designed algorithms and integrates them into the LoD2 building model, which remains static during this reconstruction phase. Consequently, our evaluation compares the ground truth with the LLM-driven refinement of the SAM~3 segmentation masks.

\subsection{Facade Feature Distance}

For evaluating façade reconstruction, metrics such as FRDS and mIoU are inherently limited because they operate strictly at the pixel level, measuring overlap between predicted and ground-truth masks. While effective for assessing segmentation accuracy, they are highly sensitive to small spatial misalignments and fail to capture higher-level properties such as semantic consistency or structural coherence. 

In contrast, a high-level feature distance evaluates similarity in a learned latent feature space, where representations encode rich semantic and contextual information. In transformer-based vision models such as DINOv2, image representations can be compared either through the CLS token (classification token), which serves as a global representation of the entire image, or through patch-level features \citep{oquab2023dinov2}. Using the CLS token, the metric reflects global semantic similarity, allowing it to assess whether the reconstructed façade preserves the correct architectural elements and overall scene meaning, even under geometric variations. Alternatively, using patch-level features enables comparison of local structure and spatial layout, capturing how well architectural elements, e.g., windows and doors, are arranged. This makes high-level feature distance metrics significantly better suited for façade reconstruction, where object consistency and structural fidelity are more important than exact pixel-wise correspondence. Let $f_{\mathrm{CLS}}(x) \in \mathbb{R}^{d}$ denote the CLS embedding of an image $x$, where $d$ is the feature dimension. The global semantic distance between two images $x_{1}$ and $x_{2}$ can then be defined as the cosine distance between their CLS embeddings:
\begin{equation}
d_{\mathrm{global}}(x_{1}, x_{2})
=
1 -
\frac{f_{\mathrm{CLS}}(x_{1})^{\top} f_{\mathrm{CLS}}(x_{2})}
{\lVert f_{\mathrm{CLS}}(x_{1}) \rVert_{2}\,\lVert f_{\mathrm{CLS}}(x_{2}) \rVert_{2}}.
\end{equation}
Subsequently, let $f_{i}(x) \in \mathbb{R}^{d}$ denote the feature embedding of patch $i$ for image $x$, with $i = 1, \dots, N$, where $N$ is the total number of image patches and $d$ is the feature dimension. The patch-wise distance between two images $x_{1}$ and $x_{2}$ can then be defined as the mean cosine distance over all corresponding patch embeddings:
\begin{equation}
d_{\mathrm{patch}}(x_{1}, x_{2})
=
\frac{1}{N}
\sum_{i=1}^{N}
\left(
1 -
\frac{f_{i}(x_{1})^{\top} f_{i}(x_{2})}
{\lVert f_{i}(x_{1}) \rVert_{2}\,\lVert f_{i}(x_{2}) \rVert_{2}}
\right).
\end{equation}
As a result, we propose a new distance metric for evaluating reconstructed façades, termed \textit{Facade Feature Distance (FFD)}, which is based on high-level feature similarity measured via cosine distance. FFD is defined as:
\begin{equation}
    \mathrm{FFD}(x_{1}, x_{2}) =
    \frac{d_{\mathrm{global}}(x_{1}, x_{2}) + d_{\mathrm{patch}}(x_{1}, x_{2})}{2},
\end{equation}
thereby combining both the preservation of the overall architectural semantics and the accurate reconstruction of the local structural layout. In other words, the metric jointly captures whether the correct architectural elements are present and how well these elements are spatially arranged. Note that the resulting distance lies in the range $[0,1]$, where values lower indicate higher similarity.

\section{Experimental Results}

To evaluate the effectiveness of our approach, we conducted a series of experiments using the previously introduced metrics, complemented by a qualitative analysis of a set of architecturally heterogeneous buildings in Hamburg, Germany.

First, we computed the mIoU between the segmentation masks produced by SAM~3 and the color- and class-adapted ground-truth masks from the eTRIMS dataset. This evaluation was intended to isolate and assess the pure semantic segmentation capability of our zero-shot approach. The resulting mIoU of 0.7221 indicates strong segmentation performance on heterogeneous building façades across Europe, as represented in the eTRIMS dataset, particularly considering that the model was neither fine-tuned nor specifically trained for this task. A substantial portion of the observed error in our experimental results can be attributed to the fact that dormer windows, which are not reconstructed by our approach, are present in the eTRIMS ground truth, as illustrated in Figure~\ref{fig:gt-normalisation}.

To evaluate the reconstruction performance with respect to the semantic and topological dimensions, we computed both the FRDS and the FFD between the LLM-refined façade masks and the corresponding color- and class-adapted ground-truth masks from the eTRIMS dataset. The FRDS is 0.7654, indicating only limited additional expressive power compared with the previously reported mIoU, as it effectively evaluates only the semantic dimension. In contrast, the FFD is 0.4945, with a $d_{global}$ of 0.4523 and a $d_{patch}$ of 0.5367, indicating that our approach performs substantially better in reconstructing façade semantics, that is, the individual façade elements, than in preserving layout or topological consistency. This provides deeper insight into the performance of reconstruction approaches than previous metrics allow. A noticeable share of the topological error can be attributed to the uncalibrated nature of the SVI data and the resulting propagation of uncertainties, which led to errors in the scale and geometric distortion of façade openings in the reconstructed LoD3 model.

\begin{figure}[ht!]
    \centering
    \includegraphics[width=\linewidth]{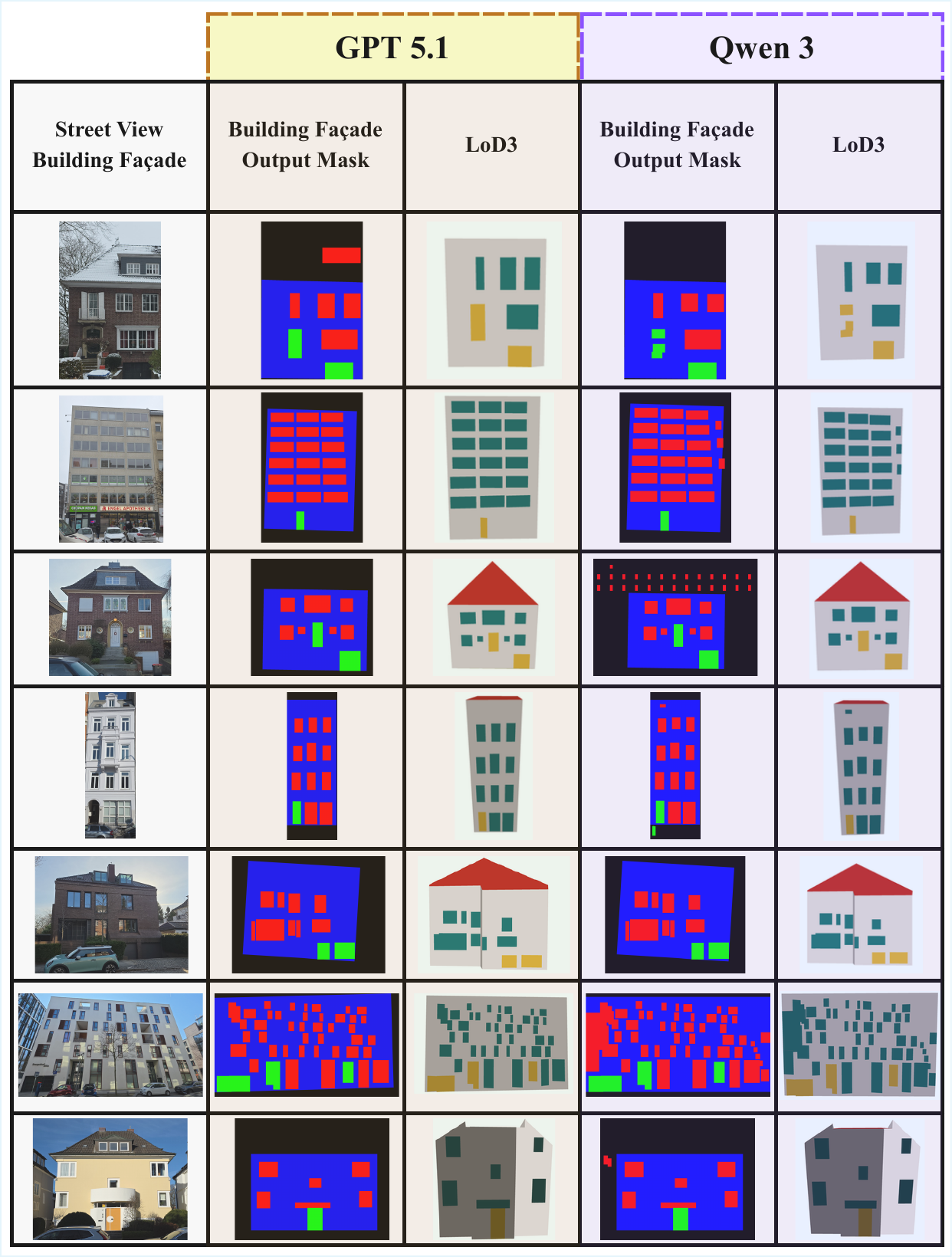}
    \caption{Exemplary results of our approach using either GPT-5.1 or Qwen~3. In the shape completed mask, façades represented in blue, windows in red, and doors in green. In the reconstructed LoD3 model, façades represented in gray, windows in green, and doors in yellow.}
    \label{fig:gpt-vs-qwen}
\end{figure}


Although the agent is guided through the denoising, shape-completion, as well as reprojection steps, and is explicitly instructed to ignore dormer windows, the approach occasionally still predicts such windows in the refined masks. 
A qualitative comparison between GPT~5.1 and Qwen~3 indicates that GPT~5.1 produces more accurate shape completion, as illustrated in the first and second last row of Figure~\ref{fig:gpt-vs-qwen}. Furthermore, GPT~5.1 exhibits fewer hallucinations than Qwen~3. This is evident in the third row, where Qwen~3 introduces several non-existent windows in the roof area, and in the fourth row, where it hallucinates an additional basement door, likely influenced by the car window visible in the foreground of the street-view image. Overall, these observations suggest that Qwen~3 is more susceptible to visual noise present in the input images or from inaccuracies in the object segmentation produced by SAM~3 in the preceding step. However, because the expert-designed reconstruction programs provided to the agent do not model dormer windows, these erroneous predictions do not further degrade the final façade reconstruction results. Interestingly, the agent is nevertheless able to detect and successfully reconstruct the shape of garage doors and small basement windows, even when these elements are partially occluded, as illustrated in the third top row of Figure~\ref{fig:gpt-vs-qwen}. Furthermore, both models are capable of identifying small windows partially hidden by façade ornamentation and balconies, as shown in the last row of Figure~\ref{fig:gpt-vs-qwen}.

To analyze the completeness of the reconstructed façades using the eTRIMS dataset, we counted the number of correctly identified windows and doors relative to the ground truth annotations. For windows, our approach achieved a detection rate of \(76.66\%\), indicating that \(23.34\%\) of the annotated windows were missed. For doors, the detection rate was \(106\%\), meaning that our approach identified more doors than were annotated in the ground truth. This discrepancy can likely be attributed to the fact that the eTRIMS ground truth contains incomplete and inaccurate annotations, e.g., for garage doors.

\section{Outlook \& Conclusion}

In this paper, we present an end-to-end pipeline for the LoD3 reconstruction of façade openings in 3D city models in a CityGML-conform format using an agent-driven system. Our approach advances existing methods by introducing a zero-shot segmentation strategy that achieves sufficiently strong results in our benchmark on the eTRIMS dataset, while eliminating the need for manual, labor-intensive annotation of training data required by supervised semantic segmentation approaches. A further distinguishing feature of our method is the agent-driven automatic validation step, which produces ready-to-use, CityGML-conform LoD3 building models.

In addition, this paper introduces a novel evaluation approach for façade reconstruction based on the reprojected segmentation masks that are also available in other methods addressing this task. The proposed metric, termed \textit{Facade Feature Distance (FFD)}, leverages high-level feature distances extracted from a vision transformer model to assess similarity in feature space rather than through pixel-wise overlap, as done by existing metrics such as FRDS or mIoU. The rationale behind FFD is that it provides a balanced assessment of both semantic correctness and architectural layout fidelity, which are equally important for façade reconstruction, instead of focusing primarily on semantic overlap as in conventional pixel-wise metrics.

Future research will focus on the semantic enrichment of additional façade characteristics, such as the reconstruction of balconies and protrusions and dormer windows, which could further strengthen the use of semantic 3D city models for urban simulation and analysis. Another promising direction is the integration of additional data sources to support more advanced reasoning strategies and to improve GPS-based map matching, thereby simplifying the correspondence between façade surfaces and camera images.

\section{Acknowledgment}

The authors gratefully acknowledge the computing time granted by the KISSKI project. This research was supported by the project ‘Next Generation City Networking’ (Grant No. 19DZ24004) at the Hanseatic Wireless Innovation Competence Center (HAWICC). The project is funded by the Federal Ministry of Transport via the German Center for Future Mobility.

{
	\begin{spacing}{1.17}
		\normalsize
		\bibliography{ISPRSguidelines_authors} 
	\end{spacing}
}

\end{document}